# Improved Squeaky Wheel Optimisation for Driver Scheduling

Uwe Aickelin, Edmund K. Burke and Jingpeng Li*

School of Computer Science and Information Technology
The University of Nottingham
Nottingham, NG8 1BB
United Kingdom
{uxa, ekb, jpl}@cs.nott.ac.uk

**Abstract.** This paper presents a technique called Improved Squeaky Wheel Optimisation (ISWO) for driver scheduling problems. It improves the original Squeaky Wheel Optimisation's (SWO) effectiveness and execution speed by incorporating two additional steps of *Selection* and *Mutation* which implement evolution within a single solution. In the ISWO, a cycle of *Analysis-Selection-Mutation-Prioritization-Construction* continues until stopping conditions are reached. The *Analysis* step first computes the fitness of a current solution to identify troublesome components. The *Selection* step then discards these troublesome components probabilistically by using the fitness measure, and the *Mutation* step follows to further discard a small number of components at random. After the above steps, an input solution becomes partial and thus the resulting partial solution needs to be repaired. The repair is carried out by using the *Prioritization* step to first produce priorities that determine an order by which the following *Construction* step then schedules the remaining components. Therefore, the optimisation in the ISWO is achieved by solution disruption, iterative improvement and an iterative constructive repair process performed. Encouraging experimental results are reported.

# 1 Introduction

Personnel scheduling problems have been addressed by mangers, operational researchers and computer scientists over the past forty years. During this period, there has been a wealth of literature on automated personnel scheduling including several survey papers that generalise the problem classification and the associated approaches (Burke et al., 2004; Ernst et al., 2004).

In brief, personnel scheduling is the problem of assigning staff members to shifts or duties over a scheduling period (typically a week or a month) so that certain constraints (organizational and personal) are satisfied. The scheduling process normally consists of two stages: the first stage involves determining how many staff must be employed in order to meet the service demand; the second stage involves allocating individual staff members to shifts and then assigning duties to individuals for each shift. Throughout the process, all industrial regulations associated with the relevant workplace agreements must be complied with.

Since personnel scheduling problems are general NP-hard combinatorial problems (Garey and Johnson, 1979) which are unlikely to be solved optimally in polynomial time, various methods such as local search-based heuristics (Li and Kwan, 2005), knowledge based systems (Scott and Simpson, 1998) and hyper-heuristics (Burke, Kendall, and Soubeiga, 2003) have been studied. Over the last few years, meta-heuristics have attracted the most attention. Genetic Algorithms (GAs) form an important class of meta-heuristics (Aickelin 2002), and have been extensively applied to personnel scheduling problems (Aickelin and Dowsland 2000 & 2003; Aickelin and White, 2004; Easton and Mansour, 1999 Li and Kwan, 2003; Wren and Wren, 1995). A number of attempts have also been made using other meta-heuristics (Shen and Kwan, 2001; Aickelin and Li 2006). The methods and techniques that have been used over the years to tackle personnel scheduling problems have tended to draw on problem-specific information and particular heuristics. In this paper, we are trying to deal with the goal of developing more general personnel scheduling systems, i.e. a method which is not designed with one particular problem in mind, but is instead applicable to a range of problems and domains.

The work that is presented here is based on the observation that, in most real world problems, the solutions consist of components which are intricately woven together. Each solution component, e.g. a shift pattern assigned to a particular employee, may be a strong candidate in its own right, but it also has to fit well with other components. To deal with these components, Joslin and Clements (1999) proposed a technique called Squeaky Wheel Optimisation (SWO), and claimed it could be a general approach for various combinatorial optimisation problems. In this paper, we analyse the limitations of the original SWO and revise it by incorporating some evolutionary features into the searching process. We term the revised version the improved SWO (ISWO). Its general idea is to break a solution down into its components and assign a score to each by an evaluation function working under dynamic environments. The scores are employed in two ways: first as fitness values which determine the chances for the components to survive in the current solution, and then they are sorted to obtain an order in which a greedy algorithm reschedules deleted components.

## 2 A General Description of the ISWO

SWO belongs to the class of non-systematic search techniques. In SWO, a priority ordering of problem components is given to a greedy algorithm that constructs a solution. That solution is then analyzed to find trouble spots, i.e. those components that are not handled as well as they could be, relative to some lower bound. The priority of the components that are trouble spots is then increased. All components, sorted in the new priority ordering are then given to the greedy constructor, with the likely result that those components will be handled better in the next solution. This construct-analyze-prioritize cycle continues until a stopping condition is reached. Joslin and Clements (1999) applied this technique on production line scheduling problems and graph colouring problems with some satisfactory results. Burke and Newall (2004) developed an adaptive heuristic framework for examination timetabling problems which was based on SWO. A hybridisation of this method (Burke and Newall, 2002) with an exam timetabling methodology based upon the Great Deluge algorithm was shown to be effective on benchmark problems (Burke et al 2004).

In essence, SWO finds good quality solutions quickly by searching in two spaces simultaneously: the traditional solution space and the new priority space. Hence it avoids many problems that other local search methods often encounter. These features allow SWO to effectively make large coherent moves to escape from unpromising regions in the search space. The construct-analyze-prioritize loop learns as it executes: problem components that are hard to handle tend to rise in the priority queue, and components that are easy to handle tend to sink.

Although SWO has achieved success in certain problems of realistic size, there exist two limitations which restrict its wider applications in domains with large problem sizes, such as many practical scheduling and rostering problems. The first limitation lies in its scalability, which is caused by SWO's construction step using greedy algorithms to construct a solution from scratch at each iteration. If the construction process could start from partial solutions which contain information of past solutions, the optimisation process would speed up significantly.

The second limitation lies in its aspect of convergence: although SWO has the ability to make large coherent moves, it is, however, poor at making small tuning moves in the solution space. Ironically, this weakness is caused by its feature of operating on dual search spaces (a "strength"). Compared to the solution of the previous iteration, a small change in the sequence of components generated by the *Prioritization* step may correspond to a large change in the corresponding solution generated by the *Construction* step. For example, moving a component forward in the sequence can significantly change its state in the actual solution, because any components occurring after it in the sequence must accommodate that component's state. However, if it was possible to restrict changes of components to the trouble-makers, e.g. by delaying part of the sequence without going through the full *Analysis* and *Prioritization* cycle, then the changes in the corresponding solutions would be relatively small.

To address the above two issues, this paper presents a new technique called ISWO, which incorporates two additional steps of *Selection* and *Mutation* into the loop. These two steps enable the ISWO to implement search by simulating an evolutionary process on a single solution. Each component in the solution has to continuously demonstrate its worthiness to stay in the solution. Hence in each iteration, a number of components will be deemed not worth keeping. The evolutionary strategy adopted may also throw out, with a low probability, some worthy components. Any deleted component is then rescheduled by using a greedy algorithm one at a time, in the order they occur in the priority sequence. Of key importance is that the admittance of a new component is analyzed by a dynamic evaluation function, which takes account of how well the prospective component will fit in with others already in the solution. The above processes are iterated together with the remainder of the classical SWO. Thus the global optimisation procedure is based on solution disruption and iterative improvement, while a constructive process is performed within.

As outlined, our proposed algorithm operates a sequence of *Analysis*, *Selection*, *Mutation*, *Prioritization* and *Construction* steps in a loop on one solution. Besides these five steps, some input parameters (e.g. stopping conditions) and a valid starting solution are initialized. In the *Analysis* step, the fitness of each component in the current solution is computed. By analyzing a solution, well-fitting and ill-fitting components can be identified. The fitness measure is then used probabilistically to select components to be discarded in the *Selection* step. Components with high fitness have a lower probability of being discarded. To get out of local optima in the solution space, it is necessary to incorporate the ability to make for uphill moves. This is achieved by the *Mutation* step which probabilistically discards even superior components of the solution.

After the above steps, a previously complete solution becomes partial due to the removal of some components, and thus the resulting partial solution needs to be repaired. Before making any repairs, the *Prioritization* step uses the results of the *Analysis* step to create priorities that in turn determine the scheduling order for the recently removed components. In this step, the previous sequence of 'trouble' components (i.e. recently removed ones) is modified: problem components with lower fitness values (i.e. more trouble-making ones) are moved towards the front of the sequence: The lower the value, the further the component is moved towards the front of the sequence. Finally, the *Construction* step repairs a broken solution by applying a greedy algorithm to reschedule the removed components, in the order that they appear in the component sequence produced by the *Prioritization*. Throughout the iteration, the best solution is retained and finally presented as the final solution.

## 3 ISWO for Driver Scheduling

### 3.1 Problem description

Bus and rail driver scheduling is represents process of partitioning blocks of work, each of which is serviced by one vehicle, into a set of legal driver shifts. The main objectives are to minimize the total number of shifts and the total shift costs. This problem has attracted much interest since the 1960's. Wren and Rousseau (1995) gave an overview of the main approaches, many of which have been reported in a series of international workshop conferences, e.g. (Voß and Daduna, 2001).

To clarify the problem, we start by introducing some terminologies used in driver scheduling (Li and Kwan, 2003). A *Relief Opportunity* (RO) is a time and place where a driver can leave the current vehicle, for reasons such as taking a meal-break, or transferring to another vehicle. The work between two consecutive ROs on the same vehicle is called *a piece of work*. The work that a single driver carries out in a day is called *a shift*, which is composed of several *spells* of work. A *spell* contains a number of consecutive pieces of work on the same vehicle, and a *schedule* is a solution that contains a set of shifts that cover all the required work. The subsequent packaging of work for actual drivers is usually performed on a weekly basis, allowing for rest days and taking into account issues such as fairness and safety regulations.

The driver scheduling problem can be formulated as a set covering integer linear programming problem: all the legal potential shifts are first constructed by heuristics that are usually highly parameterized to reflect on the driver work rules of individual companies, and then a least cost subset covering all the work is selected to form a solution schedule. In practice, the model has been extended to cater for other practical objectives and constraints (Fores et al., 2002). A typical problem may have a solution schedule requiring over 100 shifts chosen from a potential set of about 50,000.

**3.2 Implementation**

This section details how to apply the ISWO to the driver scheduling problem. Based on our problem-specific knowledge, we first set up five criteria to evaluate the structure of a shift from different aspects. Since each criterion bears some degree of uncertainty, we characterize them as individual fuzzy membership functions and aggregated these membership functions together by the way of fuzzy evaluation. The resulting aggregated function is used in a general evaluation function to analyze the fitness of each solution component (i.e. shift), and then incorporated into a constructing heuristic to enable shift selection. The steps of *Analysis*, *Selection, Mutation*, *Prioritization* and *Construction* are executed in a loop to improve a given initial solution iteratively. During each iteration, an unfit portion of the working schedule is removed. Broken schedules are repaired by the constructing heuristic. Throughout the iterations, the best is retained and finally returned as the preserved solution.

**3.2.1 Analysis**
The first *Analysis* step is to evaluate the current arrangement for each shift in a schedule. In this step, the fitness of the individual shift in a complete schedule is computed. The purpose of computing this measure is to determine, besides the structural fitness of shifts, which shifts are in positions that lead to less overlapping work time, and which shifts contribute unnecessarily to large amounts of overlapping work time. Hence we can formulate a normalized evaluation function as

$$F(S_j) = f_1(S_j) \times f_2(S_j), \ \forall j \in J \tag{1}$$

where $S_j$ denotes the shift contained in the current schedule $J$ with an index number $j$, $0 \leq f_1(S_j) \leq 1$ is the structural coefficient of shift $S_j$, and $0 \leq f_2(S_j) \leq 1$ is the overcover penalty which reflects the coverage status for shift $S_j$.

*1) Structural coefficient*
Five fuzzified criteria $u_i$ ($i= 1 ,…, 5$), characterized by associated membership functions, have been abstracted for the evaluation of the shift structure (Li 2002): Total work-time $u_1$, the ratio $u_2$ of total work-time to spreadover (i.e. the paid hours for a driver from sign on to sign off), the number of pieces of work $u_3$, the number of spells $u_4$ contained in a shift, and the fractional cover $u_5$ which is given by a linear programming relaxation. Since the evaluations by individual criteria refer to the local features of each criterion, an overall evaluation (i.e. the calculation of the structural coefficient $f_1(S_j)$ for shift $S_j$) could be made by the aggregation of these five criteria as

$$f_1(S_j) = \sum_{i=1}^{5} w_i \mu_{\tilde{A}_i}, \forall j \in \{1,...,n\} \quad (2)$$

where $\tilde{A}_i$ is the fuzzy subset on the *i*-th criterion and $w_i$ is the weight of criterion $u_i$, s.t.

$$\sum_{i=1}^{5} w_i = 1, w_i \geq 0 \quad (3)$$

The design of the membership functions for these five criteria can be briefly described as follows. Since the fitness of shift $S_j$ generally increases with the total work-time, ratio of total work-time to spreadover and number of pieces of work, respectively, the membership function $\mu_{\tilde{A}_i}$ ($i$ = 1, 2, 3) for these three factors takes the same form as

$$\mu_{\tilde{A}_i} = \begin{cases} 2\left(\dfrac{x_i - b_i}{a_i - b_i}\right)^2, & a_i \leq x_i < \dfrac{a_i + b_i}{2} \\ 1 - 2\left(\dfrac{x_i - a_i}{a_i - b_i}\right)^2, & \dfrac{a_i + b_i}{2} \leq x_i \leq a_i \end{cases} \quad (4)$$

where $x_1$ is the total work-time of $S_j$, $a_1$ is the maximum total work-time, $b_1$ is the minimum total work-time, $x_2$ is the ratio of total work-time to spreadover for $S_j$, $a_2$ is the maximum ratio, $b_2$ is the minimum ratio, $x_3$ is the number of pieces of work contained in $S_j$, $a_3$ is the maximum number of pieces of work and $b_3$ is the minimum number of pieces of work.

With respect to the criterion $u_4$, in most practical problems, the number of spells in a shift is limited to be at most four. 2-spell shifts are generally more effective than others, and 3-spell shifts are more desirable than 1-spell or 4-spell shifts. Hence, the membership function $\mu_{\tilde{A}_4}$ is defined as

$$\mu_{\tilde{A}_4} = \begin{cases} 0, & \text{if } x_4 = 1 \text{ or } x_4 = 4 \\ 0.5, & \text{if } x_4 = 3 \\ 1, & \text{if } x_4 = 2 \end{cases} \quad (5)$$

where $x_4$ is the number of spells contained in $S_j$.

With respect to the last criterion $u_5$, extensive studies have shown that the fractional cover by linear programming relaxation provides some useful information about the significance of some of the shifts identified in the relaxed solution. In general, the higher the fractional value of the variable for a shift, the higher chance that it is present in the integer solution (Kwan et al., 2001). We use the following Gaussian distribution function $\mu_{\tilde{A}_5}$ to define criterion $u_5$. More details about this criterion can be found in (Li 2002).

$$\mu_{\tilde{A}_5} = \begin{cases} e^{\frac{\ln 0.01}{(a-b)^2}(x_5 - a)^2}, & \text{if } S_j \text{ is in the fractional cover} \\ 0, & \text{otherwise} \end{cases} \quad (6)$$

where $x_5$ is the fractional value of $S_j$ in the relaxed LP solution, $a$ is the maximum value in fractional cover and $b$ is the minimum value in fractional cover.

*2) Over-cover penalty*
The ratio of the overlapped work time to total work time in $S_j$, is also regarded as an important criterion, which can be formulated as over-cover penalty $0 \leq f_2(S_j) \leq 1$,

$$f_2(S_j) = \sum_{k=1}^{|S_j|}(\alpha_{jk} \times \beta_{jk}) \bigg/ \sum_{k=1}^{|S_j|} \beta_{jk}, \; \forall j \in J \tag{7}$$

where $|S_j|$ is the number of pieces of work in $S_j$, $\alpha_{jk}$ is 0 if work piece $k$ in $S_j$ has been covered by any other shifts $S_i$ in $J$ and 1 otherwise, and $\beta_{jk}$ is the work-time for work pieces $k$ in $S_j$.

### 3.2.2 Selection
This step is to decide whether a shift in a current schedule should be retained or discarded. The decision is made by comparing its fitness value $F(S_j)$ to $(p_s - p)$ where $p_s$ is a variable generated randomly for each iteration satisfying $0 \leq p_s \leq 1$, and $p$ is a constant no larger than 1. If $F(S_j)$ is larger than $(p_s - p)$, then $S_i$ will remain in its present allocation, otherwise $S_j$ will be removed from the current schedule. The pieces of work that $S_j$ covers are then released unless they are also covered by other remaining shifts in the schedule. By using *Selection*, shift $S_j$ with larger fitness $F(S_j)$ has higher probability to survive in the current schedule. Note that the purpose of subtracting $p$ from $p_s$ is to improve the efficiency of *Selection*. Without this operator, for example, almost all shifts in the current schedule will be removed when $p_s$ is close to 1.

### 3.2.3 Mutation
The *Mutation* step follows to mutate the retained shifts $S_j$, i.e. randomly discarding them from the partial solution at a small rate $p_m$. The pieces of work that $S_j$ covers are then released unless they are also covered by other remaining shifts in the schedule. Compared with the selection rate which is randomly generated for each iteration, the mutation rate $p_m$ should be much smaller to ensure convergence.

### 3.2.4 Prioritization
The *Prioritization* step first generates a sequence of problem shifts that need to be rescheduled (i.e. the ones that have been removed by the previous steps of *Selection* and *Mutation*). Using the results of *Analysis*, the problem shifts are sorted in ascending order of their fitness values, with poor-scheduled shifts being earlier in the sequence.

The obtained sequence of problem shifts is then used indirectly to determine the order in which a new solution is constructed. Since each shift constitutes a number of pieces of work, the sequence of shifts can be transformed into a longer sequence of pieces of work, with pieces that have already been covered by earlier shifts not appearing again. Thus, the new sequence consists of all the uncovered pieces of work, in the order that they would be covered by the construction heuristic described below.

### 3.2.5 Construction
The *Construction* task is to assign shifts to all uncovered pieces of work to repair a broken schedule. By considering all potential shifts with respect to the pieces of work to be covered, it is possible to build a coverage list for each piece containing all shifts that are able to cover it. The greedy constructor assumes that the desirability of adding shift $S_j(j)$ into the partial schedule increases with its function value $F(S_j)$. The reconstructing heuristic is to assign shifts until every piece of work is covered. Candidate shifts are then assigned to the unassigned pieces of work sequentially. The criterion of choosing the next uncovered piece of work for assignment is to locate the first piece of work appearing in the priority sequence, obtain its corresponding coverage list, and randomly select a shift with one of the $k$-largest function value $F(S_j)$. For a feasible solution obtained in such a way, over-cover is often inevitable and ultimately has to be resolved by manual editing before the schedule is implemented: In practise, the intervention is simply to decide a shift that should contain the over-covered pieces of work and then remove this piece from the other shifts.

Note that the evaluation function used in the *Constructing* heuristic takes the same form as the one used in the *Analysis* step. The major difference is that the former one needs to evaluate all unused shifts from the large possible legal shift set, for the purpose of selecting some shifts to form a feasible schedule, while the latter only evaluates shifts in the current schedule.

### 3.3 Experimental results

Among various heuristic and meta-heuristic approaches developed in recent years for driver scheduling, the Self-Adjusting Approach (SAA) performs generally best on a set of standard test problems (Li and Kwan, 2005). It uses the following weighted-sum objective function, which combines the two main objectives of minimizing total cost and number of shifts into a weighted-sum cost function:

$$\text{Minimize } \sum_{i=1}^{L}(c_{J_i} + 2000) \tag{8}$$

where $L$ is the number of shifts in the schedule, $c_{Ji}$ is the cost of the $i$-th shift, and 2000 is used to give priority to the first objective of minimizing the number of shifts.

For a benchmark comparison, the same objective function is used in the ISWO coded in C++ and implemented on a Pentium IV 2.1 GHz machine under Window XP. Thirteen real world instances from medium to very large size are used as the testbed. Starting from an initial solution generated by a genetic algorithm (Li and Kwan, 2003), we set the stopping criterion equal to 1000 iterations without further improvement. Also, we apply a fixed weight distribution of membership functions, $W=(0.20, 0.10, 0.10, 0.20, 0.40)$, in equation (3) to all thirteen data instances. In addition, we set parameter $p_s$ in Section 3.2.2 to be 0.3, the mutation rate $p_m$ in Section 3.2.3 to be 0.05, and the $k$ value in Section 3.2.5 to be 2. For each instance, we run the program ten times by using different random seeds.

Table 1 lists the comparative results of the ISWO against the results of the ILP and the SAA, respectively. It also lists the results of the original SWO, which are far from optimal. Each data instance was run ten times by fixing the parameters and varying the pseudo random number seed at the beginning. Compared with the solutions of the ILP approach, our best solutions are 0.78% better in terms of total shift numbers, and are only 0.11% more expensive in terms of total cost. However, our results are much faster in general, especially for larger cases. Compared with the SAA which outper-forms other meta-heuristics available in the literature (Li and Kwan, 2005), our ISWO performs better for all data instances using similar execution times.

| | SAA S | SAA C | SAA CPU | SWO S | SWO C | SWO CPU | ISWO S | ISWO C | ISWO CPU | ☐S % | ☐C % |
|---|---|---|---|---|---|---|---|---|---|---|---|
| B1 | 35 | 294 | 28 | 37 | 321 | >999 | 34 | 292 | 121 | 0.0 | 1.2 |
| B2 | 35 | 294 | 26 | 36 | 319 | >999 | 35 | 291 | 61 | 2.9 | 0.5 |
| B3 | 74 | 830 | 216 | 81 | 908 | >999 | 73 | 828 | 203 | -2.7 | -2.7 |
| T1 | 62 | 507 | 131 | 67 | 554 | >999 | 62 | 507 | 141 | 0.0 | 0.4 |
| T2 | 117 | 998 | 167 | 124 | 1097 | >999 | 116 | 994 | 176 | 0.0 | -0.9 |
| T3 | 51 | 406 | 11 | 56 | 455 | >999 | 50 | 403 | 19 | 0.0 | -0.2 |
| T4 | 62 | 572 | 530 | 67 | 632 | >999 | 61 | 569 | 536 | -4.7 | 1.2 |
| T5 | 243 | 2249 | 981 | 262 | 2488 | >999 | 242 | 2248 | 873 | 0.0 | 0.0 |
| T6 | 271 | 2102 | 130 | 314 | 2410 | >999 | 270 | 2082 | 135 | -2.2 | -0.0 |
| T7 | 343 | 2662 | 358 | 399 | 3091 | >999 | 342 | 2662 | 318 | -2.0 | 0.0 |
| T8 | 390 | 3239 | 986 | 447 | 3686 | >999 | 389 | 3200 | 928 | -1.5 | 2.0 |
| R1 | 49 | 420 | 23 | 53 | 444 | >999 | 49 | 420 | 27 | 0.0 | 0.0 |
| R2 | 49 | 414 | 59 | 54 | 437 | >999 | 49 | 411 | 74 | 0.0 | 0.6 |
| **M** | **137** | **1153** | **280** | **154** | **1295** | **>999** | **136** | **1147** | **278** | **-0.8** | **0.1** |

**Table 1.** Comparative results. B – Bus, T – Train, R – Tram, **M – Mean**. *S – best shift*, *C – best cost*, *CPU – mean CPU time in seconds*. SAA – Self-Adjusting Approach, SWO - Squeaky Wheel Optimisation, ISWO – Improved Squeaky Wheel Optimisation, TRACS – TRACS II by Fores et al. The last two columns show % between ISWO and TRACS II.

## 4 Conclusions

This paper presents a new technique to solve personnel scheduling problems by using the original idea of SWO but by adding two steps of *Selection* and *Mutation* into its loop of *Analysis / Prioritization / Construction*. With these two additional steps, the drawbacks of the original SWO in terms of optimisation ability and execution speed are successfully dealt with. Taken as a whole, the ISWO implements evolution on a single solution and carries out search by solution disruption, iterative improvement and an iterative constructive process. The experiments have demonstrated that the ISWO performs very efficiently and competitively. In general, it outperforms the previous best-performing approaches reported in the literature.

The architecture of the ISWO is innovative, and thus there is still some room for further improvement. For example, we currently only use one fixed rule. We believe that by adding some more flexible rules into the search, solution quality could be improved further. This would be particularly interesting if we have more difficult instances to solve. In the future, we are also looking at more advanced methods of *Analysis*, *Selection* and *Mutation*.

## Acknowledgements

The research described in this paper was funded by the Engineering and Physical Sciences Research Council (EPSRC), under grant GR/S70197/1.